\documentclass{article}
\usepackage{spconf,amssymb,amsmath,graphicx,multicol}
\graphicspath{{imgs}}

\title{Optimized Learned Image Compression for Facial Expression Recognition\vspace{-5pt}}

\name{Xiumei Li$^{\star}$, Marc Windsheimer$^{\star}$, Misha Sadeghi$^{\dagger}$, Björn Eskofier$^{\dagger\ddagger}$, André Kaup$^{\star}$ \thanks{The authors gratefully acknowledge that this work has been funded by the Deutsche Forschungsgemeinschaft (DFG, German Research Foundation) under project number 426084215.}\vspace{-5pt}}
\address{\begin{tabular}{cc}$^{\star}$ Multimedia Communications and Signal Processing & $^{\dagger}$ Machine Learning and Data Analytics Lab\\ Friedrich-Alexander-Universität & Friedrich-Alexander-Universität\\ Cauerstr. 7, 91058 Erlangen & Carl-Thiersch-Str. 2b, 91052 Erlangen\end{tabular}\vspace{2pt}\\$^{\ddagger}$ Translational Digital Health Group, Institute of AI for Health\\ Helmholtz Zentrum München - German Research Center for Environmental Health\\Ingolstädter Landstr. 1, 85764 Neuherberg\vspace{-2pt}}

\begin{document}
%
\maketitle
\begin{abstract}

Efficient data compression is crucial for the storage and transmission of visual data. However, in facial expression recognition (FER) tasks, lossy compression often leads to feature degradation and reduced accuracy. To address these challenges, this study proposes an end-to-end model designed to preserve critical features and enhance both compression and recognition performance. A custom loss function is introduced to optimize the model, tailored to balance compression and recognition performance effectively. This study also examines the influence of varying loss term weights on this balance. Experimental results indicate that fine-tuning the compression model alone improves classification accuracy by \(0.71~\%\) and compression efficiency by \(49.32~\%\), while joint optimization achieves significant gains of \(4.04~\%\) in accuracy and \(89.12~\%\) in efficiency.  Moreover, the findings demonstrate that the jointly optimized classification model maintains high accuracy on both compressed and uncompressed data, while the compression model reliably preserves image details, even at high compression rates.

\end{abstract}
\vspace{-7pt}

\section{Introduction}
\label{sec:intro}
\vspace{-2pt}
Facial expressions are a vital form of non-verbal communication in human emotional interactions, capable of conveying a wide range of emotions such as happiness, sadness, and anger~\cite{ekman1971constants}. 
FER is a technology designed to automatically analyze and classify facial expressions into emotional states through computational methods. In recent years, with the rapid development of deep learning, FER has achieved significant improvements in accuracy. As a critical human-computer interaction technology, FER has demonstrated broad application potential in various fields, including social robotics, healthcare, and driver fatigue monitoring~\cite{li2020deep}.

In the application of FER technology, image compression is indispensable for enabling the efficient storage and transmission of facial expression data. Compression techniques address the challenges of limited channel bandwidth and memory capacity by reducing the size of the data. Information in an image can generally be categorized into three types: useful, redundant, and irrelevant~\cite{jamil2023learning}. Through compression, irrelevant information can be discarded, allowing critical details to be emphasized. For images with redundant information, it is essential to extract and preserve the most relevant features while eliminating unnecessary content~\cite{piran2020multimedia}.

Although existing standard lossy compression methods, such as JPEG and HEVC~\cite{sullivan2012overview}, as well as recent deep learning-based image compression methods have achieved significant progress in compression efficiency, they often fail to account for the performance requirements of FER tasks. At high compression rates, these methods tend to compromise critical semantic features and lose facial details, leading to a significant decline in the accuracy of FER~\cite{xiao2022identity}.

To address the aforementioned challenges, this study proposes an end-to-end framework that integrates a compression module with a deep FER model to optimize the process of facial expression data compression. A custom-designed loss function is employed to measure the distance between the compressed data and the original data in the facial expression feature space. Extensive experimental results demonstrate that the proposed end-to-end framework not only effectively preserves critical facial expression features but also significantly improves data compression efficiency while reducing the degradation of FER performance caused by excessive compression.

\section{Related Work}
\label{sec:format}
\begin{figure*}[!t]
    \centering
    \includegraphics[width=0.9\textwidth]{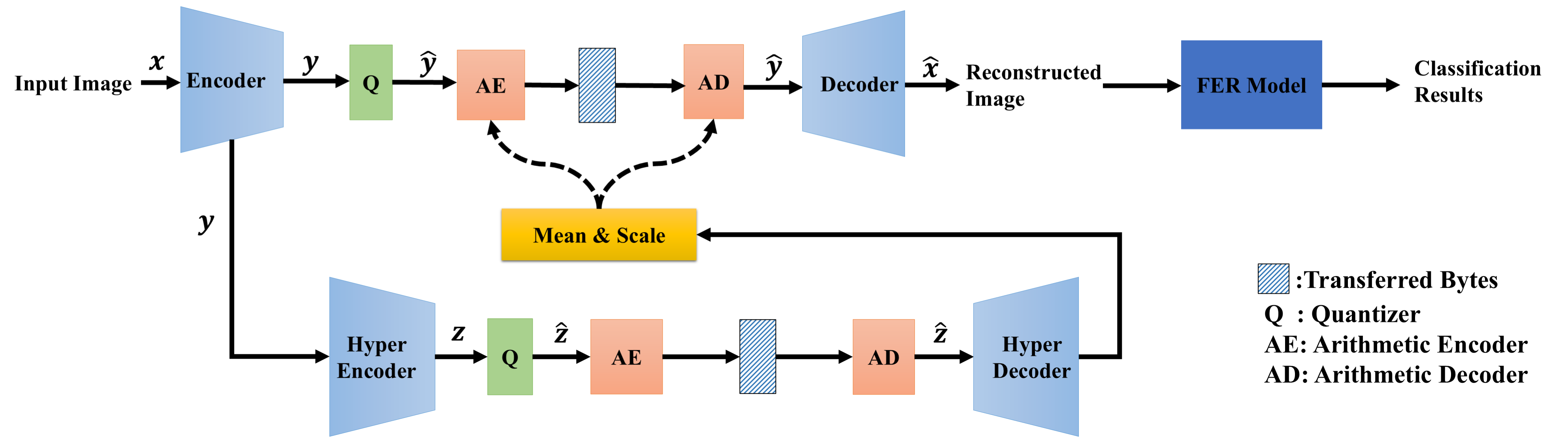} 
    \vspace{-10pt} 
    \caption{End-to-End Compression and Classification Framework for FER.}
    \label{fig:end_to_end_framework}
    \vspace{-0.5cm}
\end{figure*}

In recent years, image compression techniques based on deep learning have made significant advancements. Using the representational power of deep neural networks, these methods effectively balance reconstruction quality and compression efficiency, achieving performance comparable to traditional standards such as HEVC and AV1~\cite{han2021a} in both visual fidelity and bitrate optimization~\cite{chamain2021end}.

In \cite{balle2018variational}, an image compression framework based on autoencoders has been introduced. This approach first utilizes an autoencoder to convert the input image into a quantized feature tensor (or latent representation), which is then compressed using entropy coding. Additionally, a hyperprior module is introduced to model the probability distribution of the latent representation, providing supplementary contextual information and significantly improving compression efficiency. This architecture has become a foundational framework in the domain of deep learning-based image compression, serving as the basis for many subsequent advancements and optimizations~\cite{minnen2018joint,minnen2020channel,cheng2020learned,mentzer2020high}.

Our research is based on~\cite{minnen2018joint}, which builds upon and extends the framework proposed by~\cite{balle2018variational}. This model enhances the probability modeling of latent features by integrating joint autoregressive and hierarchical priors. The autoregressive approach captures pixel-by-pixel dependencies within the latent representation, effectively reducing data redundancy and improving compression efficiency. Simultaneously, the hierarchical prior further optimizes the feature modeling process by capturing contextual relationships within the latent information.

Recent advances in task-specific image compression have demonstrated the effectiveness of end-to-end optimization frameworks in enhancing the performance of downstream computer vision tasks. The work~\cite{chamain2021end} proposed novel end-to-end optimized image compression frameworks that integrate compression modules with downstream vision tasks such as object detection. By jointly optimizing encoder, decoder, and task-specific networks, these methods ensure that critical semantic information is preserved during compression while maintaining the performance of downstream tasks.

This concept has been further advanced by performing classification and semantic segmentation directly on compressed representations, without decoding to pixel space~\cite{torfason2018towards}. By jointly optimizing the compression network and task-specific network, this approach significantly improves the performance of classification and segmentation tasks, particularly at low bitrates.

In~\cite{wang2019scalable}, a facial image compression framework that integrates analysis and reconstruction was introduced. It utilizes FaceNet~\cite{schroff2015facenet} to extract deep features, which are then quantized and compressed, while simultaneously reconstructing the images to enhance visual quality at low bitrates. Xiao~\textit{et al.}~\cite{xiao2022identity} combined deep learning-based image compression models with facial image recognition tasks and introduced an identity-preserving loss function. This method directly optimizes the semantic consistency of facial features during the compression process, ensuring that the compressed images retain sufficient identity information. As a result, it achieves high compression rates while minimizing the impact on facial recognition performance.

\section{End-To-End Framework}
\label{sec:pagestyle}

Unlike existing studies, this research proposes an end-to-end compression optimization model specifically designed for FER tasks. The model jointly optimizes the image compression module and the expression recognition module while incorporating a specially designed loss function. This loss function integrates bitrate loss, mean squared error (MSE) loss, and cross-entropy loss, effectively balancing the trade-off between compression rate and recognition performance. By preserving critical expression features during the compression process, the model significantly enhances recognition accuracy under high compression rates. Furthermore, this study explores the effects of different components of the loss function on compression efficiency and recognition accuracy, providing deeper insights into optimizing the balance between compression rate and facial expression recognition performance.

\subsection{End-to-End Model Architecture}

Our end-to-end model framework, as illustrated in Fig.~\ref{fig:end_to_end_framework}, consists of a learning-based image compression model and a FER model. The compression model comprises a core autoencoder and a hyper network. The core autoencoder encodes high-resolution input images into low-dimensional latent representations and reconstructs them, while the hyper network models the probabilities of quantized latent representations to improve entropy encoding efficiency.

Specifically, the encoder transforms the input image $x$ into a latent representation $y$, which is then quantized. The hyper encoder maps $y$ into a hyper-latent representation $z$, which is quantized and entropy-encoded into bitstreams. These bitstreams are decoded to obtain the Gaussian distribution's means and variances, which facilitate a more accurate understanding of the global structure and statistical distribution of the latent representation during the entropy encoding and decoding processes in the core autoencoder. Finally, the decoder reconstructs the image. To ensure differentiability during training, uniform random noise is added to the latent representations to approximate the quantization effect.

Given the end-to-end architecture of the model, it is essential to balance network complexity and the performance of the classification module. To achieve this, a lightweight ResNet-18 network~\cite{he2016deep} is utilized for the FER module. This network processes decoded images from the compression module as input and performs facial expression classification.

\subsection{Joint Loss Function}

To simultaneously satisfy the requirements of image compression and FER tasks, we design a joint loss function \(L_{\text{total}}\), which combines the rate loss \(L_{\text{rate}}\), a MSE-based distortion loss \(L_{\text{distortion}}\), and the cross-entropy loss of the FER task \(L_{\text{CE}}\) through weighted summation. By adjusting the weights of each component, we can achieve a balance between compression rate, reconstruction quality, and classification accuracy. The mathematical expression is as follows:

\begin{equation}
    L_{\text{total}} = \alpha \cdot L_{\text{rate}} + \beta \cdot L_{\text{distortion}} + \gamma \cdot L_{\text{CE}} \text{ .}
\end{equation}

Here, the rate loss \(L_{\text{rate}}\) measures the entropy of the latent representations and is defined as:
\begin{equation}
    L_{\text{rate}} = \mathbb{E}_{y \sim q(y|x)}[-\log_2 p(y)] \text{ ,}
\end{equation}
where \(y\) represents the latent representations, \(p(y)\) is the probability distribution of the latent representations modeled by the hyper network, and \(q(y|x)\) is the approximate posterior distribution of the latent representations.

The distortion loss \(L_{\text{distortion}}\) quantifies the pixel-wise difference between the original image \(x\) and the reconstructed image \(\hat{x}\), using the Mean Squared Error (MSE):

\begin{equation}
    L_{\text{distortion}} = \frac{1}{N} \sum_{i=1}^N (x_i - \hat{x}_i)^2 \text{ ,}
\end{equation}
where \(x_i\) and \(\hat{x}_i\) denote the original and reconstructed values of the \(i\)-th pixel, respectively, and \(N\) is the total number of pixels in the image.
The usage of MSE as distortion loss constraints that the reconstructed images remain similar in the pixel space to the original images.
Thus, the decoded images remain interpretable by a human observer.

The cross-entropy loss \(L_{\text{CE}}\) measures the alignment between the predicted probability distribution \(\mathbf{p}\) and the ground-truth label distribution \(\mathbf{q}\), and is defined as:

\begin{equation}
    L_{\text{CE}} = -\sum_{i=1}^C q_i \log (p_i) \text{ ,}
\end{equation}
where \(C\) denotes the total number of classification categories, and \(q_i\) and \(p_i\) represent the ground-truth and predicted probabilities for the \(i\)-th category, respectively.

During training, the model minimizes the joint loss function \(L_{\text{total}}\) to achieve collaborative optimization of both image compression and FER tasks. The weight parameters \(\alpha\), \(\beta\), and \(\gamma\) can be adjusted according to experimental requirements to explore the trade-offs between different task objectives.

\subsection{Training Strategies}

In this study, two training strategies were adopted to optimize the proposed model. The first strategy involves fixing the pre-trained weights of the classification model while fine-tuning only the weights of the compression model. This allows the FER classification model to act as supervising task loss, guiding the compression model to preserve task-relevant features during compression without affecting classification performance on uncompressed data.

The second strategy involves jointly fine-tuning both the compression and classification models, where the parameters of both models are updated simultaneously. This enables the compression model to achieve higher compression rates while preserving critical features for classification and allows the classification model to adapt to compressed image representations. Through mutual adaptation during training, this approach enables the system to achieve efficient image compression while maintaining excellent classification accuracy. 

\section{Experimental Setup}

This study utilized the AffectNet~\cite{mollahosseini2017affectnet} dataset for training and testing, which is one of the largest available facial expression datasets, containing over 1 million facial images across 8 emotional categories. The dataset used in this research consists of facial images in RGB format with a resolution of 224\texttimes224, including 287,651 images in the training set and 3,999 images in the test set.

In this study, the compression model consistently adopts the pretrained Mean Scale Hyperprior model (mbt2018-mean) \cite{minnen2018joint} with weights from the CompressAI\footnote{https://github.com/InterDigitalInc/CompressAI} library. The model is optimized for MSE and bitrate. This model is used as the baseline codec for all compression modules in this work, ensuring consistent and reliable performance.

The classification model in this study uses a pre-trained ResNet-18 with weights provided by LibreFace~\cite{chang2024libreface}. These weights were pre-trained and distilled on the EmotionNet, FFHQ, and AffectNet datasets and were comprehensively evaluated and validated on the AffectNet test set and RAF-DB dataset, demonstrating excellent generalization capability and robustness, making it well-suited for the facial expression recognition task in this work.

Both training strategies in this study utilized the same dataset and identical training parameter settings. Specifically, two optimizers were employed: AdamW as the primary optimizer and Adam as the auxiliary optimizer. The weight decay for AdamW was set to 1e-2, and the initial learning rate for both optimizers was 3e-5. The batch size was 16, dropout rate 0.1, and gradient clipping threshold 0.1. Training was conducted for up to 30 epochs with early stopping based on the validation loss. Image preprocessing included resizing to 256×256 resolution, random flipping, and normalization.

\section{Results and Discussion}

\begin{figure}[tb]
    \centering
    \includegraphics[width=0.8\linewidth]{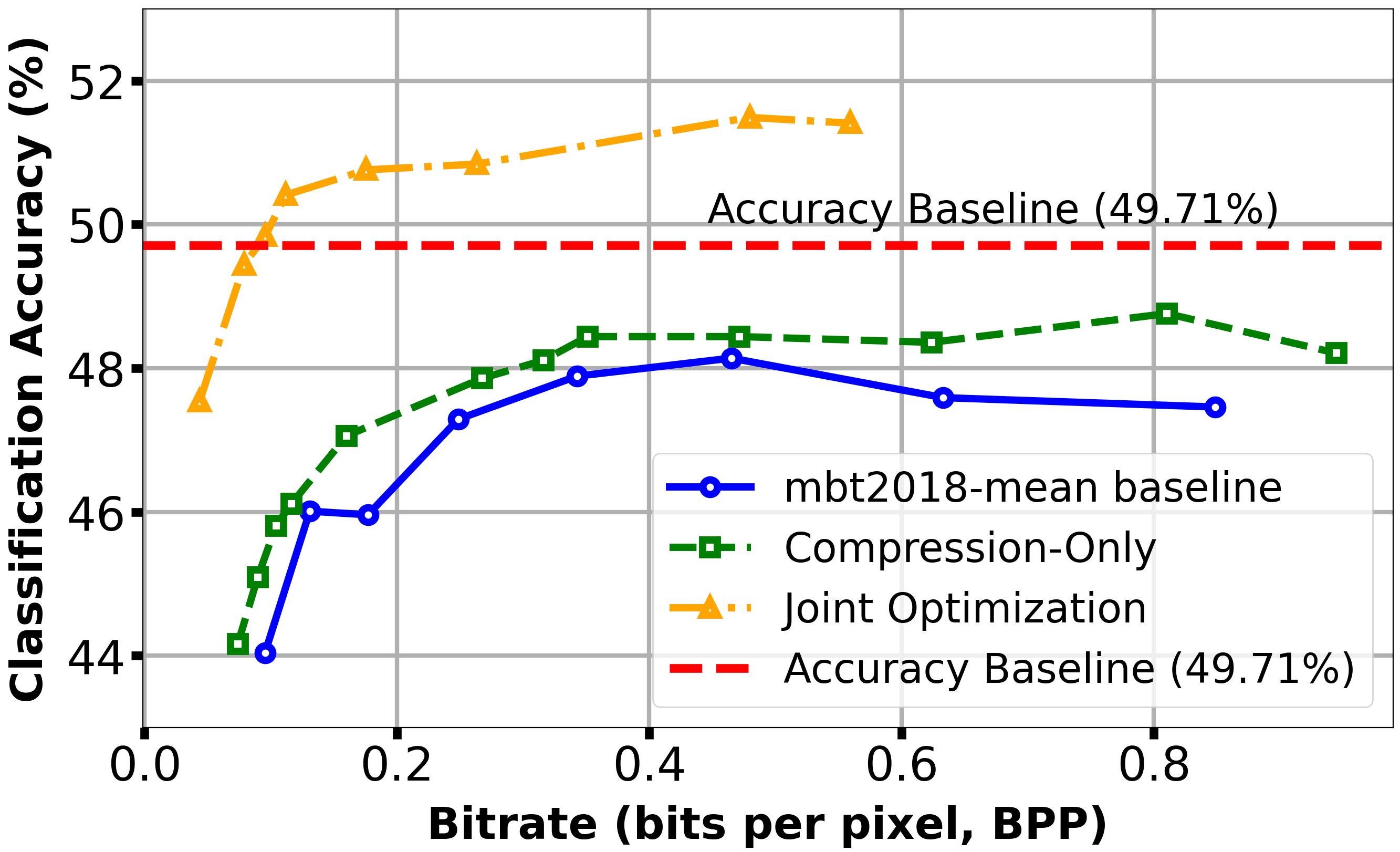} 
    \caption{Rate-Accuracy comparison for different compression methods and the base accuracy of the pretrained FER model.}
    \vspace{-0.5cm}
    \label{fig:rate_accuracy_comparison}
\end{figure}

Fig.~\ref{fig:rate_accuracy_comparison} compares classification accuracy under different compression rates for three methods. The performance evaluation uses bits per pixel (BPP) for compression ratio and Top-1 classification accuracy of the 8-class classification on the AffectNet test set for recognition tasks. The red dashed line represents the accuracy baseline, showing the Top-1 classification accuracy (49.71~\%) of the pre-trained FER model on the uncompressed AffectNet test set. The rate-accuracy curve for the \textit{mbt2018-mean} baseline represents the results obtained directly by inferring with the pretrained compression model and FER model, where images are compressed at different quality levels \(q\) and then processed by the classifier. The curve for \textit{Compression-Only} represents the end-to-end method that fine-tunes only the compression model, while \textit{Joint Optimization} shows the joint optimization of the compression model and the FER model. 

The results indicate that both end-to-end methods consistently outperform the \textit{mbt2018-mean} baseline across all bitrates, demonstrating the effectiveness of end-to-end model. Among these, the \textit{Joint Optimization} approach achieves the best overall performance, showcasing strong adaptability to varying compression levels. Notably, at low bitrates \(\sim0.1\) BPP, where traditional models often suffer from accuracy degradation, the jointly optimized model not only maintains competitive recognition accuracy but surpasses the accuracy baseline. This underscores the robustness of the joint optimization strategy in balancing compression efficiency and recognition performance.

To further quantify and compare the performance improvements of the end-to-end models under two training strategies, this study calculates the Bjontegaard Delta (BD) metrics. Specifically, BD-Accuracy represents the average improvement in classification accuracy within the same bitrate range, while BD-Rate measures the percentage improvement in compression rate at equivalent classification accuracy, both relative to the \textit{mbt2018-mean} baseline. Table~\ref{tab:bd_comparison} shows that \textit{Compression-Only} improves classification accuracy by 0.71~\% and reduces the compression rate by 49.32~\%. The \textit{Joint Optimization} achieves a 4.04~\% increase in accuracy and an 89.12~\% reduction in compression rate, highlighting its significant advantage in image compression and FER.

\begin{table}[t]
\centering
\setlength{\tabcolsep}{6pt} %
\begin{tabular}{|l|c|c|}
\hline
\textbf{Method} & \textbf{BD-Accuracy} & \textbf{BD-Rate} \\ \hline
Compression-Only & 0.71~\% & -49.32~\% \\ \hline
Joint Optimization & 4.04~\% & -89.12~\% \\ \hline
\end{tabular}
\caption{Comparison of training strategies in terms of BD-Accuracy and BD-Rate against \textit{mbt2018-mean} as baseline.}
\vspace{-0.5cm}
\label{tab:bd_comparison}
\end{table}

\begin{figure*}[tb]
\begin{minipage}[b]{0.33\linewidth}
  \centering
  \centerline{\includegraphics[width=\linewidth]{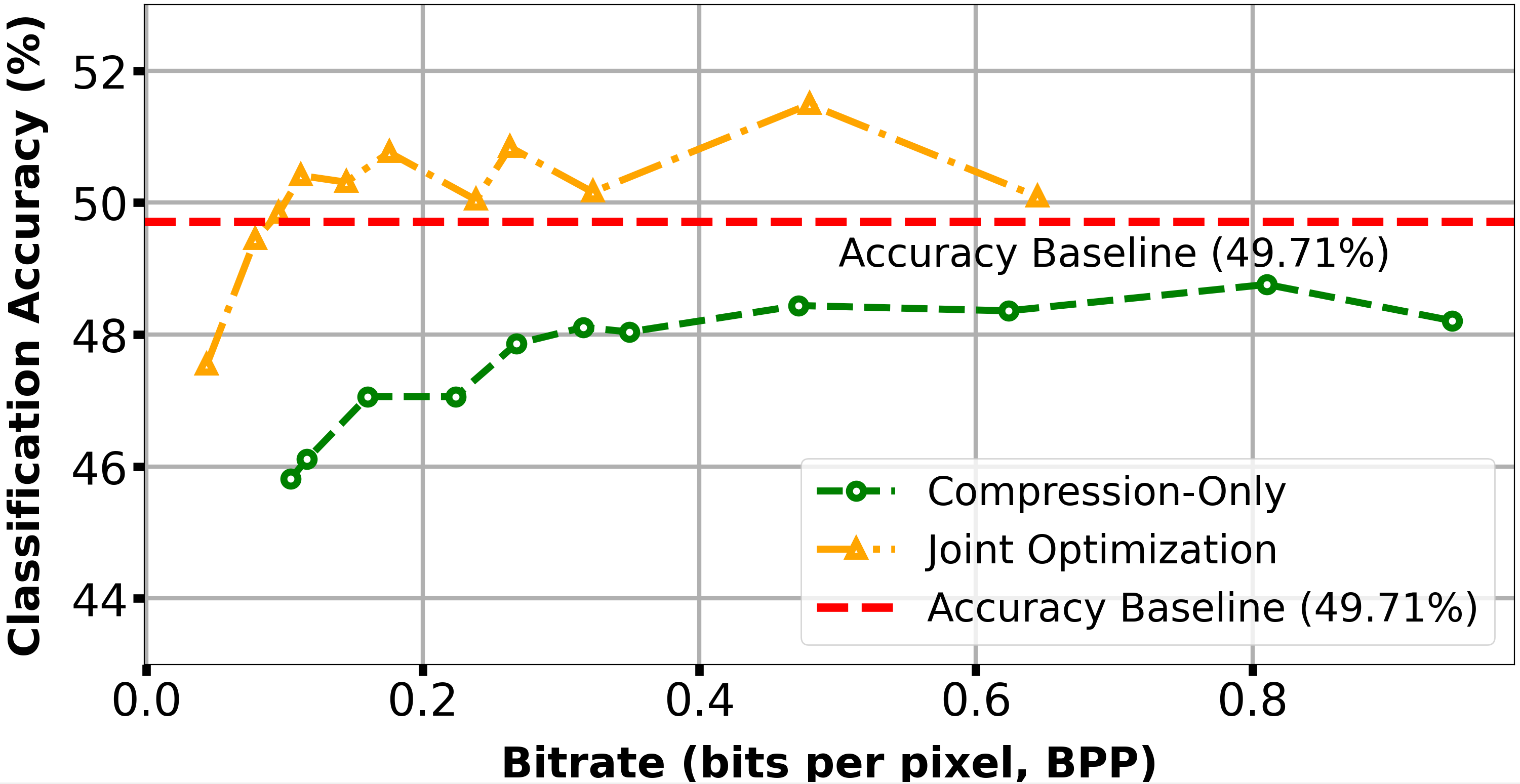}}
  \centerline{(a) Impact of BPP loss weight}\medskip
\end{minipage}
\hfill
\begin{minipage}[b]{0.33\linewidth}
  \centering
  \centerline{\includegraphics[width=\linewidth]{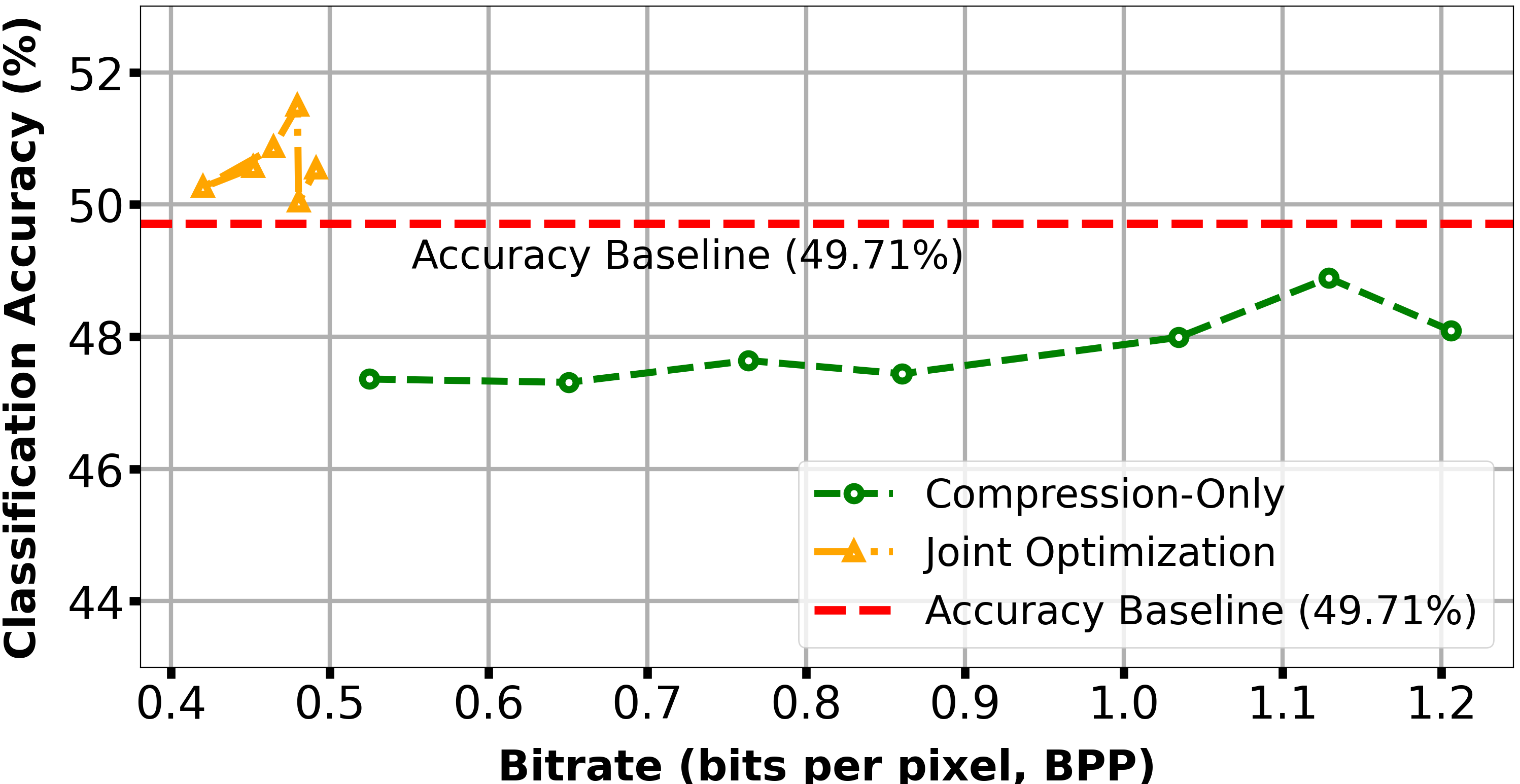}} 
  \centerline{(b) Impact of cross-entropy loss weight}\medskip
\end{minipage}
\hfill
\begin{minipage}[b]{0.33\linewidth}
  \centering
  \centerline{\includegraphics[width=\linewidth]{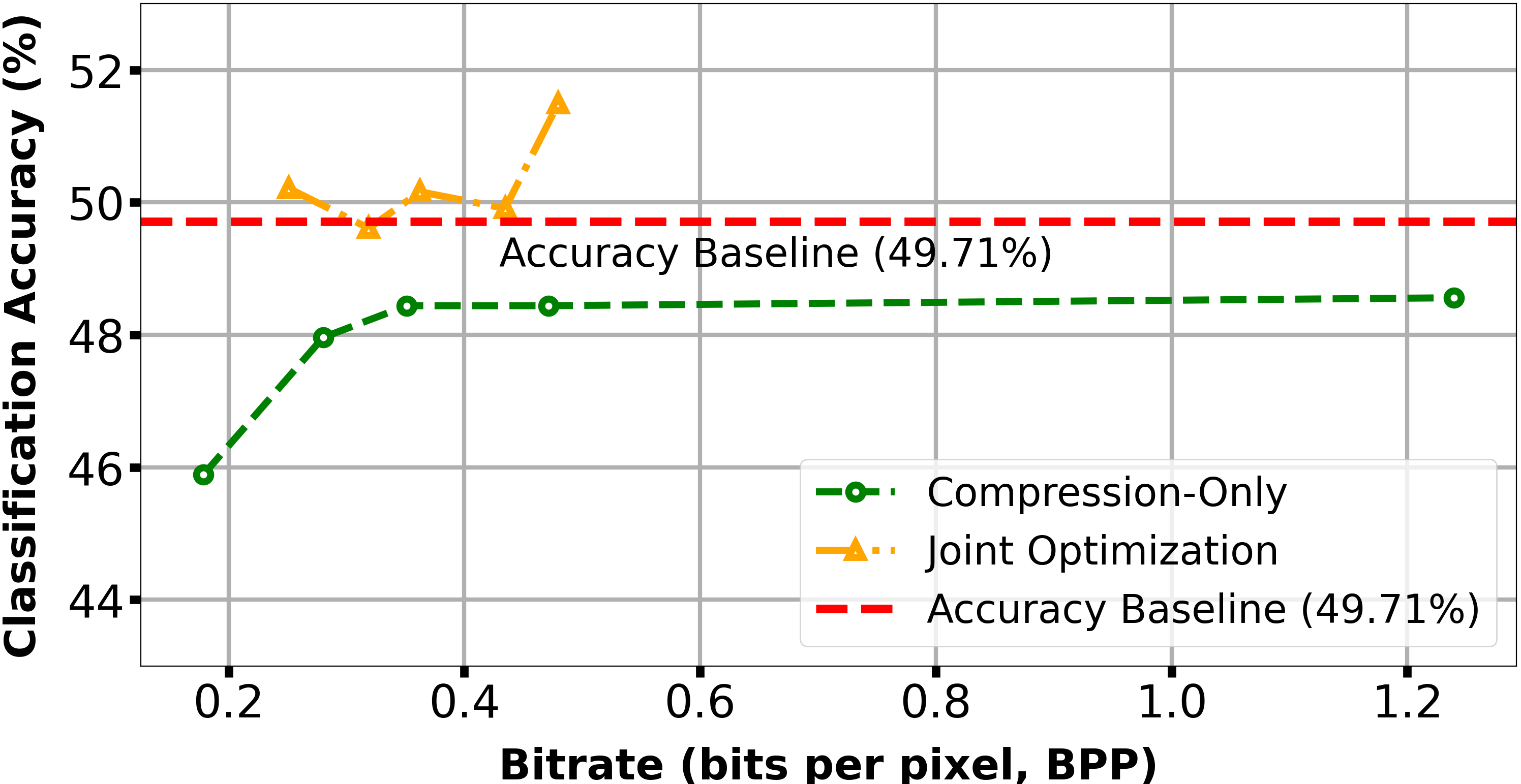}}
  \centerline{(c) Impact of MSE loss weight}\medskip
\end{minipage}
\vspace{-0.9cm} %
\caption{Impact of different loss components on the classification accuracy and compression rate under two training strategies.}

\label{fig:loss_impact}
\end{figure*}

\subsection{Impact of Loss Components}
To optimize the trade-off between compression efficiency and classification accuracy, we analyze the impact of the individual loss terms on the performance of the end-to-end model. During the experiments, each loss weight was adjusted independently while keeping the other weights fixed, enabling a precise evaluation of the effect of specific loss terms on classification accuracy and bitrate. This process identified the optimal combination of loss weights, providing valuable insights for the optimization of the end-to-end model.

Figs.~\ref{fig:loss_impact} illustrate the impact of the different losses on the performance of the end-to-end model under two training strategies. Experimental results reveal similar trends for the end-to-end models across both training strategies. As the weights of BPP loss and cross-entropy loss decrease, both bitrate and classification accuracy improve. However, excessively low BPP weights compromise compression efficiency with limited accuracy gains. Similarly, smaller cross-entropy loss weights outperform larger weights in terms of classification accuracy, contrary to the initial expectation that a high cross-entropy weight improves classification accuracy. In contrast, increasing the MSE loss slightly improves accuracy, however, excessively high MSE loss weights lead to a substantial reduction in compression efficiency.

\begin{table*}[htb]
\centering
\begin{tabular}{|c|c|c|c|c|c|c|c|c|c|}
\hline

\textbf{Bitrate (BPP)} & 0.04 & 0.08 & 0.10 & 0.11 & 0.18 & 0.26 & 0.32 & 0.48 & 0.56 \\ \hline
\textbf{Joint Optimization (\%)} & 47.54 & 49.44 & 49.84 & 50.41 & 50.76 & 50.84 & 50.16 & 51.49 & 51.41 \\ \hline
\textbf{Raw Image (\%)} & 49.34 & 50.16 & 50.24 & 50.29 & 50.81 & 49.74 & 50.59 & 51.36 & 51.49 \\ \hline
\end{tabular}
\caption{Impact of joint optimization on classification accuracy: the classification model, extracted from the jointly optimized end-to-end framework, was tested on both uncompressed AffectNet test images and compressed images at various bitrates.}
\label{tab:classification_comparison}
\vspace{-0.5cm} %
\end{table*}

\subsection{Classification Improvement through Joint Optimization}
To evaluate the impact of joint optimization on the classification model, we extracted the classification model from the jointly optimized end-to-end model at various bitrates and tested it on the uncompressed AffectNet test set. These results were compared to the classification performance on compressed images. Table~\ref{tab:classification_comparison} reveals that when the bitrate exceeds approximately 0.1 BPP, the classification accuracy on compressed and uncompressed images becomes nearly identical. Interestingly, at certain bitrates, the classification accuracy on compressed images slightly surpasses that on uncompressed images, indicating a ``feature enhancement'' effect. This outcome further validates the ability of the compression module to retain critical features under joint optimization, even at low bitrates.

\subsection{Visual Quality of Decompressed Images}

\begin{figure}[htb]
\centering
\begin{minipage}[b]{0.99\linewidth}
  \centering
  \includegraphics[width=8cm]{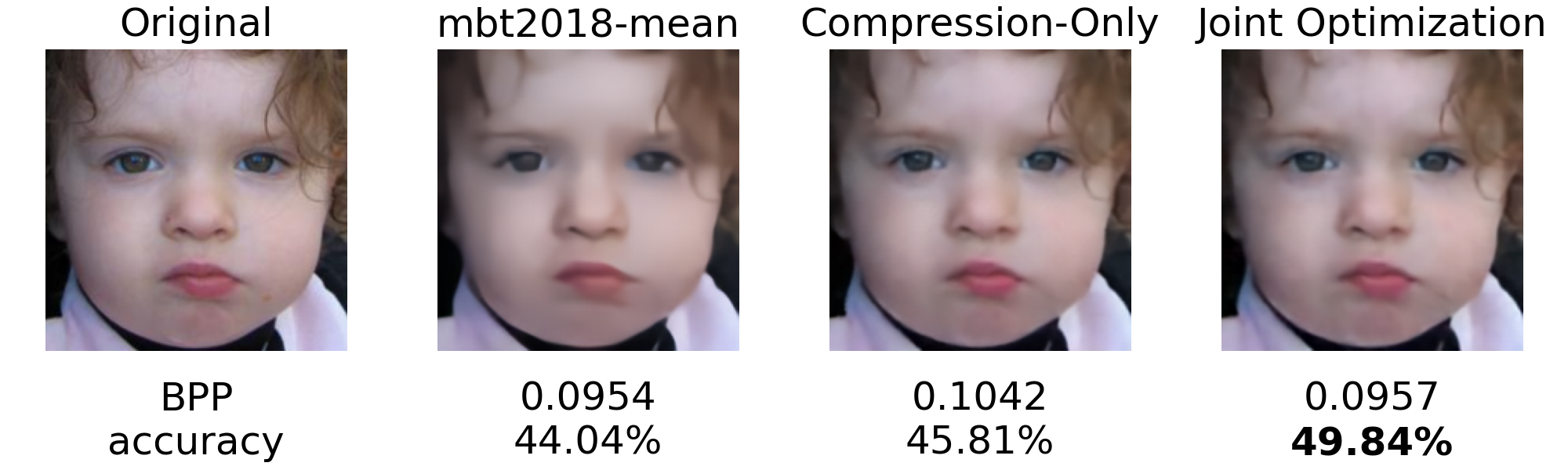}
  \centerline{(a) Low bitrate} %
\end{minipage}
\begin{minipage}[b]{0.99\linewidth}
  \centering
  \includegraphics[width=8cm]{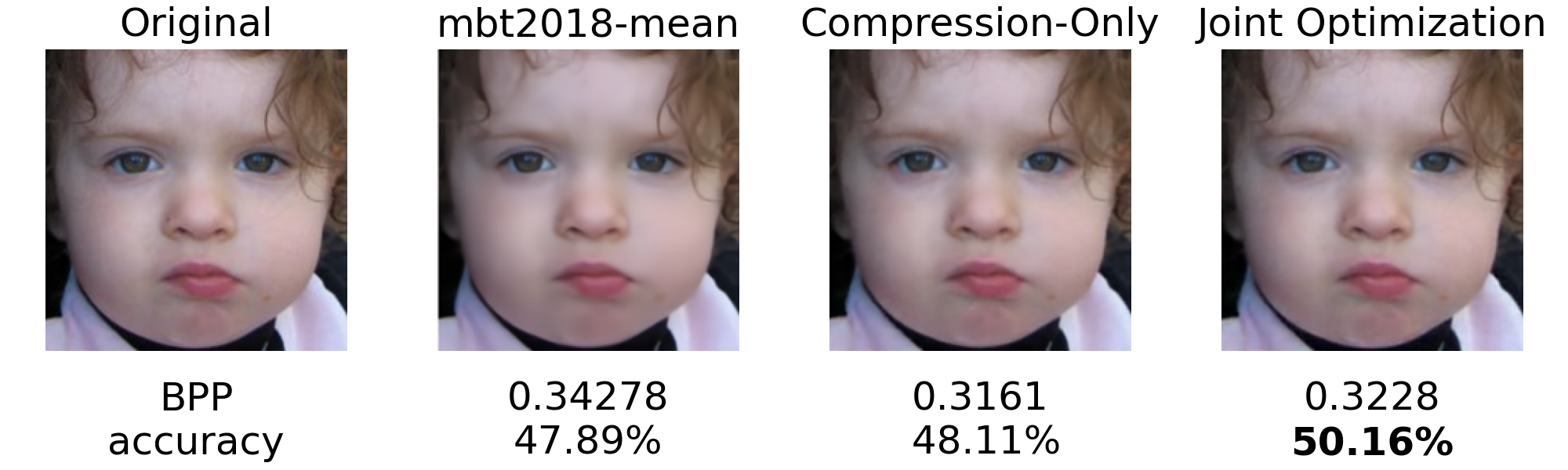}
  \centerline{(b) Medium bitrate}
\end{minipage}
\begin{minipage}[b]{0.99\linewidth}
  \centering
  \includegraphics[width=8cm]{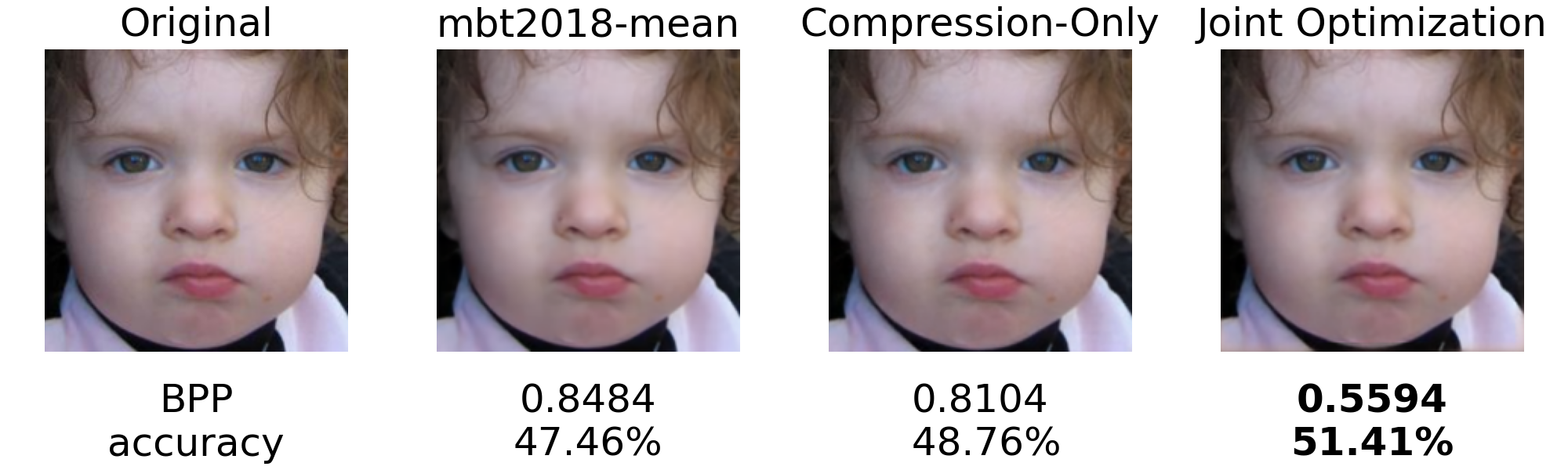}
  \vspace{-5pt}
  \centerline{(c) High bitrate}
\end{minipage}
\vspace{-3pt}
\caption{Comparison of reconstructed images for various bitrates. Our \textit{Joint Optimization} approach achieves the best rate-accuracy trade-off, without visible artifacts in the reconstructed image.}
\label{fig:comparison}
\vspace{-0.5cm} %
\end{figure}

We compare the decompressed images generated by the end-to-end models under two training strategies and the mbt2018-mean model at various compression rates in Fig.~\ref{fig:comparison}. Overall, the end-to-end models demonstrated a better ability to preserve image details at relatively low bitrates. Notably, the jointly optimized end-to-end model not only maintained higher visual clarity at low bitrates but also significantly outperformed other methods in classification tasks.

\section{Conclusion}

This study presents an end-to-end model for optimizing facial expression data compression, incorporating a custom loss function that balances bitrate, cross-entropy, and MSE. Two training strategies were evaluated, and experiments demonstrated the model’s superiority in achieving both high compression efficiency and classification accuracy across various compression rates. The analysis further highlights the impact of different components of the loss function on performance, showcasing the model’s ability to preserve image details during compression. These findings provide strong evidence for the effectiveness of end-to-end models in integrating compression and FER tasks.

Future work could investigate adaptive loss functions to balance compression and FER dynamically. Advanced architectures, such as Transformer-based models or multi-scale networks, could also be explored to enhance feature representation and improve performance in the joint optimization framework.

\vfill\pagebreak
\bibliographystyle{IEEEbib}
\bibliography{strings,refs}

\end{document}